# Machine Vision Using Cellphone Camera: A Comparison of deep networks for classifying three challenging denominations of Indian Coins

Keyur D. Joshi[1], Dhruv Shah[2], Varshil Shah[2], Nilay Gandhi[2], Sanket J. Shah[2], Sanket B. Shah[2]
[1]Asisstant Professor, [2]Undergraduate Students, Ahmedabad University, Ahmedabad, India
{keyur.joshi, dhruv.s3, varshil.s, nilay.g, sanket.s2, sanket.s1}@ahduni.edu.in

*Abstract*—Indian currency coins come in a variety of denominations. Off all the varieties ₹1, ₹2 and ₹5 have similar diameters. Majority of the coin styles in market circulation for denominations of ₹1 and ₹2 coins are nearly the same except numerals on its reverse side. If a coin is resting on its obverse side, the correct denomination is not distinguishable by humans. Therefore, it was hypothesized that a digital image of a coin resting on its either size could be classified into its correct denomination by training a deep neural network model. The digital images were generated by using cheap cell phone cameras. To find the most suitable deep neural network architecture, four were selected based on the preliminary analysis carried out for comparison. The results confirm that two of the four deep neural network models can classify correct denomination from either side of a coin with accuracy of 97%.

## I. Introduction and Background

Machine Vision usually involves object detection and classification based on available images. Coins form a major part in the Indian currency system. With the increase in computational power, coin classification is one of the emerging field of study. Since the manual classification is time consuming, tedious, and prone to errors, there is a need to have an efficient, robust and automatic coin classification system. Moreover, such systems are useful in unmanned stores with automated checkout facility [1] and securing cash circulation [2]. Indian coins have a variety of styles for all the denominations in circulation.

A number of studies on coin recognition/classification systems using neural networks are reported in literature. [3] developed a five convolutional layer-based deep neural network architecture for coin classification with 90% validation accuracy. [4] used cell phones to classify Indian currency notes with 93% validation accuracy. The authors in [5] have classified ancient Roman Republican Coins using the motifs minted on the reverse side of the coins. In [6], the authors majorly focus on the edge detection problem in coin recognition and suggest that dynamic statistical colour threshold method is the best suitable aid for the problem. Later they have used multi-channel Gabor filter for feature extraction and back propagation neural network for numeral recognition. The authors in [7] evaluated the effect on accuracy due to the design parameters which included size of the database, number of classes, quality of images, type of network, nature of training and testing strategy. A set of guidelines for part recognition tasks is presented. [8] presented a novel Multi-column Deep Neural Network for image classification. Their architecture outperformed the existing traditional models in terms of recognition accuracy on data-set such as MNIST and NORB. [9] focused on optimizing the task of Indian coins recognition at high speed for a camera-based moving coin system. [10] compared accuracy and performance of three different techniques for Indian coin classification namely particle classification, pattern matching and recognition matching. The authors reached the conclusion that none of the techniques reached the goal of 95% accuracy at 1000 coins/min. [11] proposed a coin recognition system design which consists of Robert's edge detection method, Laplacian of Gaussian edge detection method, Canny edge detection method and Multi-Level Counter Propagation Neural Network (ML-CPNN) yielding 93%, 95%, 97.25% and 99.47% recognition rate respectively.

The noise present in form of shadow or from improper positioning of coin/camera may lead to false classification. The classification accuracy is the major distinguishing performance measure for a developed classification model. However, many times a value of 100% accuracy is not good due to model's poor generalizability mainly as a result of over-fitting [12]. The motivation of this work is to develop a well generalizable classification model that can classify Indian coins from images captured using cell phones. Prior to designing such a system, a reliable database is required such that inferences can be derived and the underlying pattern can be represented in a clear manner. Indian coin denominations such as ₹ 1 and ₹ 2 have similar styles so that it is difficult to identify the denomination from a single side of the coin [10]. The absence of a good dataset for the Indian coins in this context gave rise to the need of creation of such a dataset from scratch. This work involved a database creation having Indian coins images acquired in a systematic manner.

The images in the database were processed to make it more presentable to data driven classification methods such as deep neural networks. We employ selected renowned deep learning architectures that are based on convolutional neural networks, and analyse their performance for this application.

In this paper we refer two sides of the coin as obverse and reverse, where the reverse side contains the numerical value of the coin. Section II and III provides discussion on image database creation and image processing. Section IV provides details about the deep neural networks. Results and conclusion with future work are discussed in section V and VI, respectively.

## II. Database Formation

### A. Available Database

The open source database [13], consisted 656 total images of the coins currently used in India. Specifications of the database are that it includes datasets of 194, 197, 191 and 74 images of ₹ 1, ₹ 2, ₹ 5 and ₹ 10 coins respectively. Here, the individual denominations have been segregated properly into different datasets. However, this database had a few demerits. First, the images have a lot of noise and not suitable to readily use for feature extraction. Second, background for the images was random for the entire database, reducing uniformity across the data. Third, distance between a coin and the camera used for capturing the image was non-uniform. In nutshell, a structured process has not been followed for capturing the images of the coins.

### B. Generated Database

A database was generated with structured process. Fig. 1 shows a sample setup where two views of the setup are shown. The distance between coin and the camera was kept at 10 cm after a set of experiments with higher and lower values. Three cell phones were used to capture the coin images. Table 1 lists major features of the cell phones. Features of the generated database are: 1) three different datasets to accommodate three challenging denominations of ₹ 1, ₹ 2, and ₹ 5, Fig. 2 shows only the reverse side of the coin styles considered in this work; 2) the background for the images have been prefixed to be dark and uniform without any granularity; 3) the physical coins were manually rotated with 90, 180, 270 and 360 degrees; there were total 121 physical coins used as shown in Fig. 3. Total number of images under title 'column A' is 967 considering both sides.

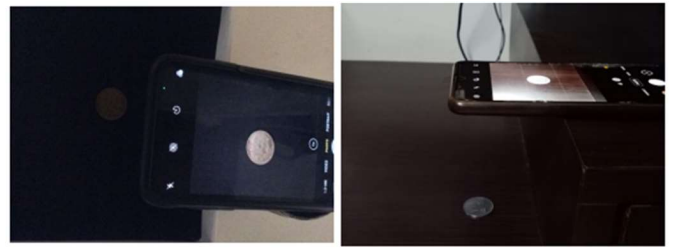

Fig. 1. Setup view from an angle (left) and from a side (right)

Table 1: major features of the three cellphones used for data acquisition

| Cellphone | Major Features | | | |
|---|---|---|---|---|
| | Model number | Shutter speed | Aperture | ISO |
| Redmi Note 9 Pro | M2003J6A1I | 1/17 second | f/1.79 | 6500 |
| Samsung Galaxy A31 | SM-A315F | 1/20 second | f/2.0 | 3200 |
| iPhone X | MQA92LL/A | 1/4 second | f/1.8 | 2000 |

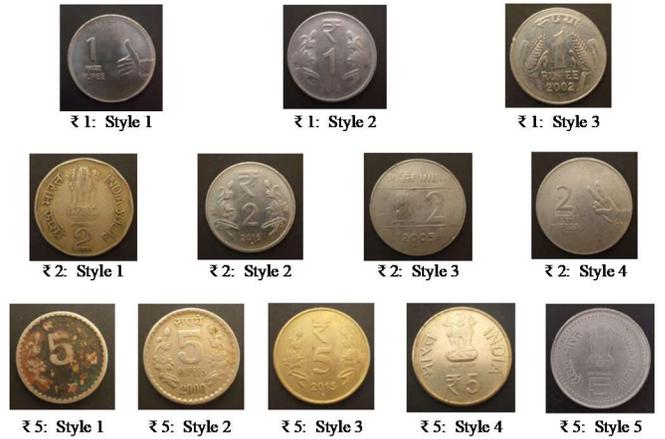

Fig. 2. Reverse side of the sample coins for ₹ 1: total 6, 20 and 17 physical coins for style 1, 2 and 3 (top); ₹ 2: total 2, 26, 5 and 8 physical coins for style 1, 2, 3 and 4 (middle); and ₹ 5: total 1, 15, 16, 3, and 1 physical coins for style 1, 2, 3, 4, and 5 (bottom). For obverse sides, refer [10].

These images are then processed as discussed in next section.

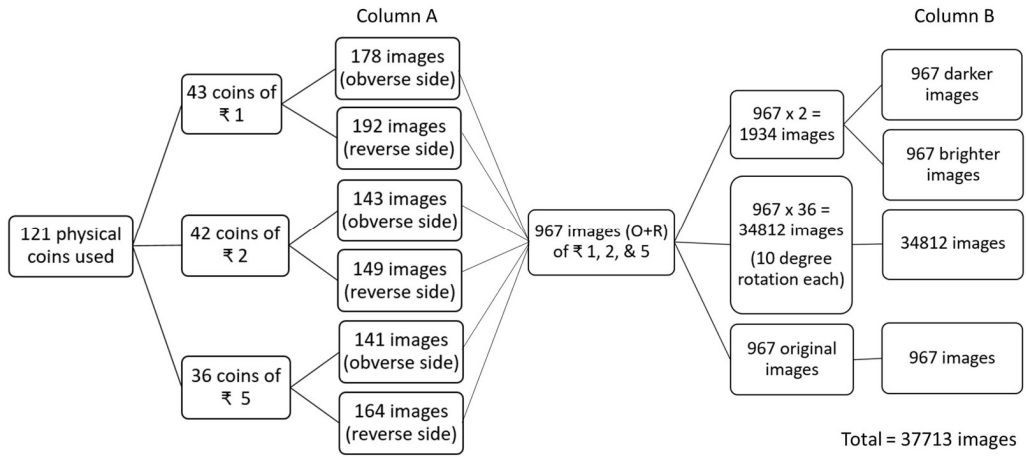

Fig. 3. Database generation: from 121 physical coins (image acquisition) to the database of 37713 images (dataset augmentation)

## III. Processing the Datasets

The images generated cannot be directly used due to reasons such as redundancy of image space and inconsistency in coin positions. Therefore these 967 images were processed such that for each one input image, there were 38 output image. Total number of images in the database including the processed input images resulted in 37713 images, shown under 'column B' in Fig. 3. The image processing involved data cleaning and augmentation.

### A. Data cleaning

Self-optimizing architectures depend on the training data for the extraction of knowledge and on the testing data to ensure the absorption of the knowledge. To speed up the procedure, removal of the redundant data is important without which the learning process may get misguided. On the other hand, important data must not be lost or the performance would suffer. Cleaning involved centering, cropping, resizing and grey-scaling. For centering and cropping, HoughCircles method provided by deep learning library OpenCV was used. The images were resized to a size of 150x150x3 and then converted to greyscale. The images were converted to greyscale because in a separate set of experiments, the color feature was not found to be significantly improving the performance.

### B. Data Augmentation

Deep Learning Models can solve complex problems as they are capable of identifying various hidden features and defining a relationship between these interdependent features. To be able to learn such intricate hidden patterns and relations, they need to identify feature themselves. This require huge amount of data. Data Augmentation synthetically increases the number of images by applying various techniques. The output images might seem similar to us, but they provide useful extra information to the machine learning model. Brightness and rotational augmentation were selected.

In brightness augmentation, the images were darkened and brightened by 33% of the original image as beyond this the change were not identifiable to a human eye. We chose to digitally rotate the original image by 10 degrees. The rotation will rotate pixels out of the image frame and leave corner areas of the frame with no pixel data that must be filled in. These corner areas were filled with the average value of all the four corner values of the original image. After applying both the augmentations, 38 new images (36 from rotation augmentation and remaining 2 from brightness augmentation) were created for each input image. The database of 37713 images was then used for training and testing deep neural networks (DNNs).

## IV. Experiments with Deep Neural Networks

The four DNN architectures were selected for performance comparison on the selected application after a set of experiments with the existing dataset [13].

### A. Selecting deep network architectures

Initially seven DNNs were tested with the database [13]: 1) AlexNet, 2) ResNet50, 3) Inception V3, 4) MobileNet V2, 5) GoogleNet, 6) VGG16 and 7) LeNet. Considering four classes and 75%-25% train test data distribution, the performance in terms of accuracy was found out using Adam optimizer with cross entropy. After 25 epochs, the training was stopped and four architectures were selected for comparative performance analysis: 1) ResNet50, 2) Inception V3, 3) MobileNet V2, and 4) VGG16.

ResNet50 has 50 layers of residual networks. This solves the problem of vanishing gradients by providing an alternative shortcut for the gradient to flow. [14]. Inception V3 is updated DNN architecture that uses aggressive regularizationa and suitably factorized convolutions for computational efficiency [15]. The VGG16 architecture is made up of very small convolutional filters and a depth of 16 weighted networks [16]. MobileNetV2 is a convolutional neural network that is designed to work on mobile devices. It is based on an inverted residual structure, with bottleneck levels connected by residual connections. The design of MobileNetV2 contains a fully convolutional layer with 32 filters as well as 19 residual bottleneck layers [17].

### B. Methodology for model development

The generated database was divided into training and testing datasets with 67%-33% distribution. The total classes prepared were 6 considering obverse and reverse side of the denomination as two separate classes. This helped in inferring which side is classification friendly. Nevertheless, it is the denomination which is important, (and not a particular side). Therefore, the six classes were merged into 3 classes for their respective observations. The performance analysis of the DNNs architectures and results are discussed in next section.

## V. Results and Discussion

The total images used for testing were 12446. Fig. 4 shows the performance plot of the test accuracy vs epochs. Here the testing accuracy was observed for up to 50 epochs. The value of testing accuracy after completion of the 30th epoch as can be seen by vertical line in Fig. 4 for all the DNN architectures under consideration seemed well generalizable. Table 2 shows the testing accuracy for obverse side, reverse side and both side combined after the end of 30th epoch. From Table 2, it is clear that InceptionNet V3 and VGG16 provided similar performance, while ResNet50 provided the less performance and MobileNet V2 provided the least performance. Let us examine the individual confusion matrices obtained using DNN architectures.

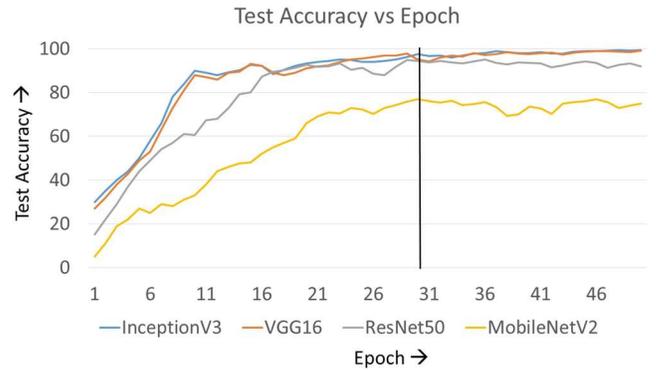

Fig. 4. Performance plot of selected four DNN architectures

Table 2: DNN architectures performance after completion of 30[th] epoch

| Deep network architectures | Test Accuracy (in %) | | | |
| --- | --- | --- | --- | --- |
| | Obverse | Reverse | Both | Both (floor) |
| MobileNet V2 | 73.78 | 70.75 | 77.03 | 77 |
| ResNet50 | 92.68 | 93.08 | 94.97 | 94 |
| Inception V3 | 96.28 | 95.92 | 97.57 | 97 |
| VGG16 | 99.81 | 99.8 | 97.87 | 97 |

A. *Results with InceptionNet V3*

Table 3 shows two confusion matrices for assessing the performance of InceptionNet V3. The left is for 6 classes and the right is for 3 classes. Misclassification between two denominations is undesirable as compared with misclassification between two sides of same denomination. Misclassifications between ₹ 1 obverse side and ₹ 2 obverse side are highlighted using red dotted oval shape. Here, ₹ 1 classified as ₹ 2, for 60 and ₹ 2 classified as ₹1 for 25 observations. This contributed majorly in misclassifications of overall confusion matrix shown in right side of Table 3. The reason behind this is high similarity between obverse sides of ₹ 1 and ₹ 2 coins. Although there were some misclassifications, the reverse side does not seem to have this problem. There were total 303 misclassifications out of 12446 observations in overall confusion matrix shown in right side of Table 3.

B. *Results with VGG16*

Table 4 shows two confusion matrix for 6 classes in left and 3 classes in right. Here the difference between classification of ₹ 1 and ₹ 2 is more pronounced as the highlighted values are larger than any other neighbor misclassifications in the confusion matrix. This architecture registered total 264 misclassifications out of 12446 observations in overall confusion matrix shown in right side of Table 4.

C. *Results with ResNet50*

Resultant confusion matrices for 6 and 3 classes obtained from ResNet50 DNN architecture are shown in left and right side of Table 5, respectively. A sum of 127 images from ₹ 2 reverse side images were classified as ₹ 1 reverse side images. Ignoring few misclassifications, this was not the case with InceptionNet V3 and VGG16. This shows ResNet50 is not performing well as compared with the previous two DNN architectures. Interestingly, 95 images of ₹ 2 obverse side were classified as ₹ 1 obverse side which is in line with the results obtained with InceptionNet V3 and VGG16. The same is highlighted in confusion matrix shown in left side of Fig. 5. The results show 625 misclassifications from 12446 observations of overall confusion matrix.

Table 3: InceptionNet V3 performance confusion matrix: 3 denominations with two sides separately considered (left) and 3 denominations (right)

| IV3 | ₹1 Rev | ₹2 Rev | ₹5 Rev | ₹1 Obv | ₹2 Obv | ₹5 Obv | Total |
|---|---|---|---|---|---|---|---|
| ₹1 Rev | 2412 | 28 | 19 | 54 | 8 | 24 | 2545 |
| ₹2 Rev | 27 | 1783 | 8 | 17 | 61 | 12 | 1908 |
| ₹5 Rev | 7 | 2 | 2032 | 3 | 0 | 66 | 2110 |
| ₹1 Obv | 18 | 2 | 2 | 2140 | 60 | 32 | 2254 |
| ₹2 Obv | 0 | 8 | 4 | 25 | 1755 | 19 | 1811 |
| ₹5 Obv | 1 | 0 | 3 | 2 | 1 | 1811 | 1818 |
| Total | 2465 | 1823 | 2068 | 2241 | 1885 | 1964 | 12446 |

| IV3 | ₹1 | ₹2 | ₹5 | Total |
|---|---|---|---|---|
| ₹1 | 4624 | 98 | 77 | 4799 |
| ₹2 | 69 | 3607 | 43 | 3719 |
| ₹5 | 13 | 3 | 3912 | 3928 |
| Total | 4706 | 3708 | 4032 | 12446 |

Table 4: VGG16 performance confusion matrix: 3 denominations with two sides separately considered (left) and 3 denominations (right)

| VGG16 | ₹1 Rev | ₹2 Rev | ₹5 Rev | ₹1 Obv | ₹2 Obv | ₹5 Obv | Total |
|---|---|---|---|---|---|---|---|
| ₹1 Rev | 2500 | 4 | 0 | 0 | 0 | 0 | 2504 |
| ₹2 Rev | 36 | 1893 | 0 | 1 | 0 | 0 | 1930 |
| ₹5 Rev | 11 | 19 | 2031 | 0 | 0 | 0 | 2061 |
| ₹1 Obv | 31 | 1 | 0 | 2088 | 137 | 7 | 2264 |
| ₹2 Obv | 10 | 3 | 0 | 31 | 1847 | 0 | 1891 |
| ₹5 Obv | 3 | 1 | 1 | 3 | 0 | 1788 | 1796 |
| Total | 2591 | 1921 | 2032 | 2123 | 1984 | 1795 | 12446 |

| VGG16 | ₹1 | ₹2 | ₹5 | Total |
|---|---|---|---|---|
| ₹1 | 4619 | 142 | 7 | 4768 |
| ₹2 | 78 | 3743 | 0 | 3821 |
| ₹5 | 17 | 20 | 3820 | 3857 |
| Total | 4714 | 3905 | 3827 | 12446 |

Table 5: ResNet50 performance confusion matrix: 3 denominations with two sides separately considered (left) and 3 denominations (right)

| RN50 | ₹1 Rev | ₹2 Rev | ₹5 Rev | ₹1 Obv | ₹2 Obv | ₹5 Obv | Total |
|---|---|---|---|---|---|---|---|
| ₹1 Rev | 2373 | 57 | 18 | 46 | 8 | 2 | 2504 |
| ₹2 Rev | 127 | 1703 | 37 | 28 | 34 | 1 | 1930 |
| ₹5 Rev | 62 | 4 | 1962 | 4 | 0 | 29 | 2061 |
| ₹1 Obv | 103 | 15 | 12 | 2074 | 43 | 17 | 2264 |
| ₹2 Obv | 26 | 105 | 6 | 95 | 1648 | 11 | 1891 |
| ₹5 Obv | 15 | 3 | 135 | 27 | 7 | 1609 | 1796 |
| Total | 2706 | 1887 | 2170 | 2274 | 1740 | 1669 | 12446 |

| RN50 | ₹1 | ₹2 | ₹5 | Total |
|---|---|---|---|---|
| ₹1 | 4596 | 123 | 49 | 4768 |
| ₹2 | 276 | 3490 | 55 | 3821 |
| ₹5 | 108 | 14 | 3735 | 3857 |
| Total | 4980 | 3627 | 3839 | 12446 |

Table 6: MobileNet V2 performance confusion matrix: 3 denominations with two sides separately considered (left) and 3 denominations (right)

| MNV2 | ₹1 Rev | ₹2 Rev | ₹5 Rev | ₹1 Obv | ₹2 Obv | ₹5 Obv | Total |
|---|---|---|---|---|---|---|---|
| ₹1 Rev | 1365 | 361 | 126 | 496 | 136 | 61 | 2545 |
| ₹2 Rev | 152 | 1100 | 112 | 103 | 389 | 74 | 1930 |
| ₹5 Rev | 84 | 163 | 1365 | 44 | 103 | 351 | 2110 |
| ₹1 Obv | 395 | 233 | 96 | 1191 | 255 | 84 | 2254 |
| ₹2 Obv | 74 | 370 | 68 | 113 | 1103 | 98 | 1826 |
| ₹5 Obv | 49 | 99 | 384 | 55 | 116 | 1078 | 1781 |
| Total | 2119 | 2326 | 2151 | 2002 | 2102 | 1746 | 12446 |

| MNV2 | ₹1 | ₹2 | ₹5 | Total |
|---|---|---|---|---|
| ₹1 | 3447 | 985 | 367 | 4799 |
| ₹2 | 442 | 2962 | 352 | 3756 |
| ₹5 | 232 | 481 | 3178 | 3891 |
| Total | 4121 | 4428 | 3897 | 12446 |

*D. Results with MobileNet V2*

Table 6 shows two confusion matrix as per the convention followed by previous three figures. Two highlighted values in dotted red oval shape represents misclassification between ₹ 1 and ₹ 2 both sides. In particular, 361 ₹ 1 reverse side images were classified as ₹ 2 reverse side images, while 255 ₹ 1 obverse side images were classified as ₹ 2 obverse side images. Here, the problem with reverse side is more pronounced as compared with ResNet50. This problem was almost non-existent in performance of VGG16 and InceptionNet V3 DNN architectures. Left confusion matrix in Table 6 shows more misclassifications such as 496 for ₹ 1, but is of less importance as they are between two sides of the same denomination. Total of 2859 misclassifications registered out of 12446 observations in overall confusion matrix shown in right of Table 6.

*E. Results comparison*

Table 2 suggests that considering floored accuracy value for DNN architecture comparison, InceptionNet V3 and VGG16 provided similar performance. While VGG16 has estimated 138 million trainable parameters, InceptionNet V3 has roughly 5 million trainable parameters. Therefore, in terms of less computational head for this application, InceptionNet V3 is the most suitable DNN architecture. ResNet50 has roughly 23 million parameters, but was unable to provide performance close to the ones shown by Inception V3 and VGG16. MobileNet V2 performance was not up to the mark although it was developed for the applications involving mobile vision such as this one.

## VI. CONCLUSION AND FUTURE WORK

For the application of three challenging denominations of Indian coins classification, four DNN architectures were compared: Inception V3, ResNet50, VGG16 and MobileNet V2. The computationally efficient and higher performance architecture was InceptionNet V3. MobileNet V2 could not provide the satisfactory performance. There is a plan to develop cell phone application to identify the challenging denominations of Indian coins and investigate MobileNet V2 further to make it readily adoptable for this application in future.

ACKNOWLEDGMENT

The authors acknowledge the help and support received from Mr. Nisarg Patel, Mr. Devarsh Suthar and Mr. Tejas Chauhan for this research work.